\documentclass[10pt,twocolumn,letterpaper]{article}

\usepackage{iccv}
\usepackage{times}
\usepackage{epsfig}
\usepackage{graphicx}
\usepackage{amsmath}
\usepackage{amssymb}
\usepackage{multirow}
\usepackage{subfigure}
\usepackage{stfloats}
\usepackage[noend]{algpseudocode}
\usepackage{color}
\usepackage{algorithmicx,algorithm}
\usepackage{authblk}

\usepackage[breaklinks=true,bookmarks=false]{hyperref}

\iccvfinalcopy 

\def\iccvPaperID{2024} 

\ificcvfinal\fi

\begin{document}

\title{Re-ID Driven Localization Refinement for Person Search}

\newcommand*\samethanks[1][\value{footnote}]{\footnotemark[#1]}
	\author{Chuchu Han$^{1}$\thanks{Equal contribution.}, Jiacheng Ye$^2$\samethanks, Yunshan Zhong$^{2}$, Xin Tan$^{3}$, Chi Zhang$^{4}$, Changxin Gao$^{1}$\thanks{Corresponding author: cgao@hust.edu.cn}, Nong Sang$^{1}$\\
$^{1}$Key Laboratory of Ministry of Education for Image Processing and Intelligent Control, \\
School of Artificial Intelligence and Automation, Huazhong University of Science and Technology\\
$^{2}$Peking University $^{3}$Shanghai Jiao Tong University $^{4}$Megvii Technology\\
{\tt\small {\{hcc, cgao, nsang\}@hust.edu.cn \{yejiacheng, Zhongyunshan\}@pku.edu.cn}} \\
{\tt\small {tanxin2017@sjtu.edu.cn zhangchi@megvii.com}}}

	


\maketitle
 \ificcvfinal\thispagestyle{empty}\fi
\begin{abstract}
	Person search aims at localizing and identifying a query person from a gallery of uncropped scene images. Different from person re-identification (re-ID), its performance also depends on the localization accuracy of a pedestrian detector. The state-of-the-art methods train the detector individually, and the detected bounding boxes may be sub-optimal for the following re-ID task. To alleviate this issue, we propose a re-ID driven localization refinement framework for providing the refined detection boxes for person search. Specifically, we develop a differentiable ROI transform layer to effectively transform the bounding boxes from the original images. Thus, the box coordinates can be supervised by the re-ID training other than the original detection task. With this supervision, the detector can generate more reliable bounding boxes, and the downstream re-ID model can produce more discriminative embeddings based on the refined person localizations.
Extensive experimental results on the widely used benchmarks demonstrate that our proposed method performs favorably against the state-of-the-art person search methods.
\end{abstract}
\begin{figure}[htbp]
	\small
	\begin{center}
		\includegraphics[width=0.8\linewidth]{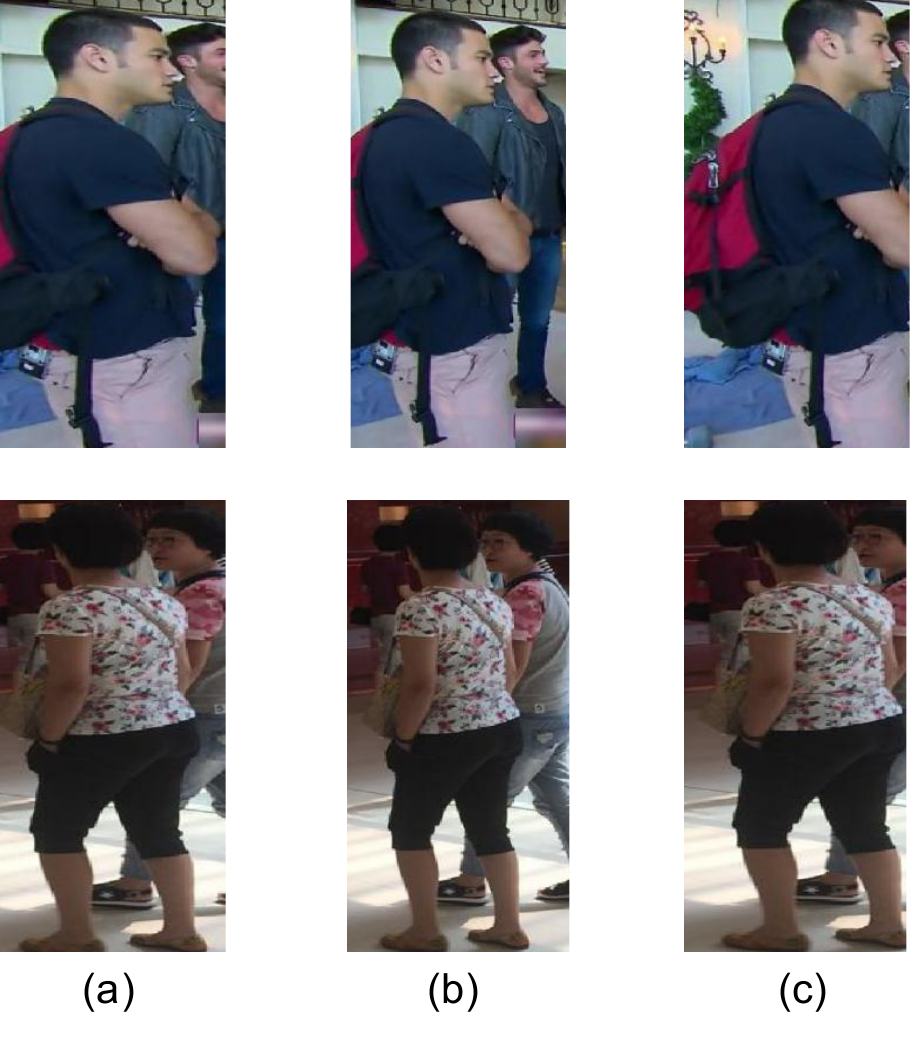}
	\end{center}
	\caption{Detection results for person search. (a) is the ground truth, (b) is detected by a general detector, and (c) is refined by the proposed method. 
		We can see that the general detector trained individually cannot favor the following re-ID well actually. It involves more noise and lacks some important details.
		By comparison, our method can refine the localization by removing excessive background and increasing useful attributes, which can produce more discriminative embeddings for the person search task.}
	\label{fig:motivation}
\end{figure}
\section{Introduction}
Although person re-ID has achieved remarkable retrieval performance~\cite{zheng2016person,chen2017beyond,wang2018mancs}, it still has major limitations for practical applications since it lacks person detection stage. 
Towards further exploring the pedestrian recognition in real-world applications, researchers have proposed the person search task~\cite{xiao2017joint, zheng2017person, chen2018person, lan2018person, Hao2017Neural, xiao2019ian} whose objective is to simultaneously localize and recognize a person from raw images. 
Instead of only matching the manually cropped or pre-detected pedestrians, person search aims to combine the task of pedestrian detection and person re-ID into a unified system. 
%

Current state-of-the-art methods~\cite{chen2018person,lan2018person} train the detector and re-ID model independently, which separate the person search task into two individual stages. However, the purpose of the detection stage is to detect the person. It cannot cover some crucial attributes (e.g., the bag) and even brings in some interferences (e.g., nearby persons), as shown in Fig.~\ref{fig:motivation}(b). Thus the incoherent framework has limitations that the detection stage cannot provide optimal proposals for the downstream person re-ID stage.
%
%
With the sub-optimal proposals, the performance of the following person re-ID stage is affected.
%
So the result of person search is also unsatisfactory.

To address the above issues, we propose a re-ID driven localization refinement network that joints the task of pedestrian detection and person re-identification in an end-to-end framework. 
Our main purpose is to optimize the detector under the supervision of re-ID loss so as to produce reliable bounding boxes. 
Specifically, we develop a differentiable ROI transform layer to realize the crop operation by affine transformations. 
Based on this layer, the bounding boxes generated by the detector can be transformed into corresponding images which are then fed to the re-ID network. 
Thus, the framework can be optimized in an end-to-end fashion.
Under the guidance of re-ID loss, the original bounding boxes will be refined for providing more reliable ones for the person re-ID. 
As shown in Fig.~\ref{fig:motivation}(c), compared with the separated detector, our method can remove interferences like nearby person and background distinctly. 
Moreover, some fine-grained details can be attended to, such as the bags, which contain discriminative information for the downstream person re-ID. 

Besides, we propose a proxy triplet loss because it is intractable to construct standard triplets in the person search pipeline.

In summary, the contributions of this paper mainly include:
\begin{itemize}
	\vspace{-1mm}
	\item We propose an end-to-end localization refinement framework for person search. Owing to the supervision of re-ID loss, more reliable bounding boxes can be produced by the detector for person search task.
	\vspace{-2mm}
	\item We introduce a differentiable ROI transform layer for the purpose of cropping the bounding boxes from the raw images.
	\vspace{-2mm}
	\item The performance on two benchmarks CUHK-SYSU and PRW achieves 93.0\% and 42.9\% on mAP, respectively, which outperforms state-of-the-art methods by a large margin. 
\end{itemize}

\section{Related Works}
\paragraph{Pedestrian Detection.}
There are several commonly used detectors in traditional pedestrian detection, including Deformable Part Model (DPM)~\cite{felzenszwalb2010object}, Aggregated Channel Features (ACF)~\cite{dollar2014fast}, Locally Decorrelated Channel Features (LDCF)~\cite{nam2014local} and Integrate Channel Features (ICF)~\cite{dollar2009integral}.
In recent years, with the development of CNN, quite a few CNN-based methods have been emerged, which can be roughly divided into the one stage~\cite{lin2017focal,redmon2016you,liu2016ssd} and two-stage manners~\cite{girshick2015fast,ren2015faster,dai2016r,he2017mask}. The main difference is whether to generate the proposals. In one stage manners, Lin~\textit{et~al.}~\cite{lin2017focal} propose the RetinaNet with a focal loss for solving the problem of class-imbalance. In two-stage manners, the prominent representation is the Faster R-CNN~\cite{ren2015faster}, which proposes a region proposal network (RPN). It greatly reduces the amount of computation while shares the characteristics of the backbone network. Lin~\textit{et~al.}~\cite{lin2017feature} design a top-down architecture with lateral connections for building multi-level semantic feature maps at multiple scales, which is called Feature Pyramid Networks (FPN). Using FPN in a basic detection network can assist in detecting objects at different scales. 

Our detection network is based on the framework of Faster R-CNN+FPN with some improvement of the structures, making the detection more reliable for pedestrians rather than universal objects. 
\vspace{-3mm}
\paragraph{Person Re-Identification.}
Recently deep learning dominates the re-ID research community with significant advantages in retrieval accuracy. Most methods~\cite{zhang2017alignedreid,sun2018beyond,chen2018group,su2017pose,zhao2017deeply,li2018harmonious} aim at producing robust and discriminative image representations. Sun~\textit{et~al.}~\cite{sun2018beyond} propose the Part-based Convolutional Baseline (PCB) to extract several body parts features. Si~\textit{et~al.}~\cite{si2018dual} utilize the attention mechanism to focus on the most discriminative features. Meanwhile, some deep metric learning methods~\cite{hermans2017defense,chen2017beyond} are also widely used in person re-ID. Hermans~\textit{et~al.}~\cite{hermans2017defense} develop a triplet hard loss, which applies an online triplet hard negative mining method in a mini-batch. It can promote the result increasingly. Chen~\textit{et~al.}~\cite{chen2017beyond} propose a quadruplet loss based on triplet loss, which aims at further reducing the intra-class variations and enlarging the inter-class variations. 

In this paper, we propose a novel proxy triplet loss, which solves the problem that standard triplet loss cannot be constructed in the person search pipeline.
\vspace{-3mm}
\paragraph{Person search.}
Person search is a recently developed task~\cite{xiao2017joint,zheng2017person,xiao2019ian,chen2018person,lan2018person,chang2018rcaa}, which aims at matching a specific person among a great number of whole scene images. In the literature, there are two approaches to deal with the problem. Some works~\cite{xiao2017joint,xiao2019ian} train the detection and re-ID model in an end-to-end manner. Xiao~\textit{et~al.}~\cite{xiao2017joint} employ the Faster R-CNN as a backbone network, and share the base layers between detection and re-ID. Meanwhile, an Online Instance Matching (OIM) is proposed, which enables better convergence in the classification task with large but sparse identities. Xiao~\textit{et~al.}~\cite{xiao2019ian} apply the center loss to this task, enhancing the feature discriminative power. Other methods~\cite{chen2018person,lan2018person} train the detector and re-ID model separately. Chen~\textit{et~al.}~\cite{chen2018person} argue that the detection task pays attention to the commonness of people, while the re-ID task focuses on the differences among people. So it is unreasonable to share the features between two tasks. By training detector and re-ID model separately, better performance is obtained. The accuracy is further enhanced by modeling the foreground and original images patches in two subnets. Lan~\textit{et~al.}~\cite{lan2018person} propose a Cross-Level Semantic Alignment (CLSA) network to solve the multi-scale matching problem. After the detection, Kullback-Leibler divergence is calculated across different layers to supervise the re-ID model. Chan~\textit{et~al.}~\cite{chang2018rcaa} introduce the reinforcement learning to detection network by constantly trying to adjust the bounding box in various ways to find the perfect match.

Unlike the aforementioned methods, 
we joint the two tasks in an end-to-end framework without sharing features.

\begin{figure*}[htbp]
	\small
	\begin{center}
		\includegraphics[width=0.8\linewidth]{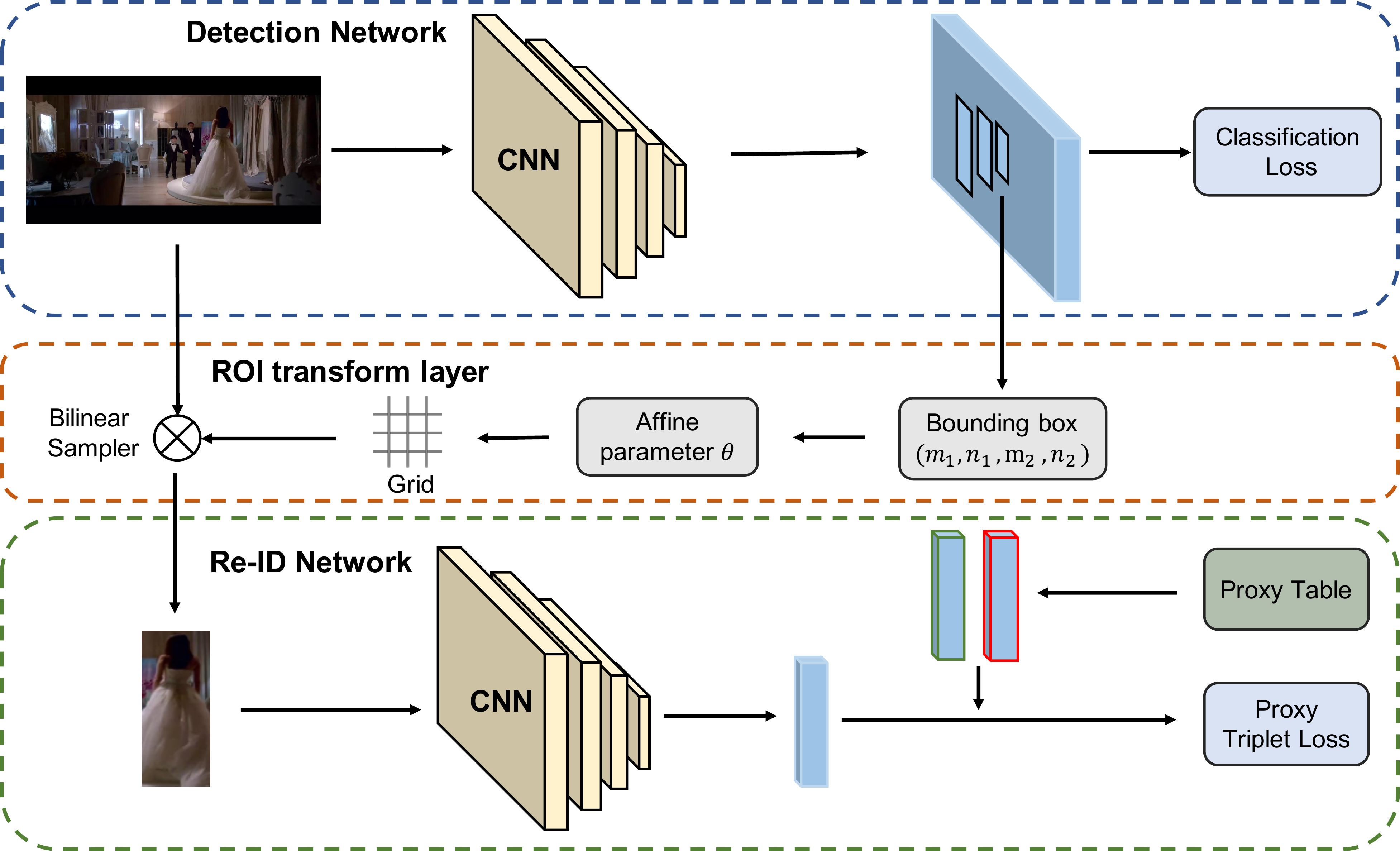}
	\end{center}
	\caption{The framework of our method. The bounding boxes generated by the detection network are sent to an ROI transform layer for the cropping operation in the raw image. Then the re-ID network extracts the features and generates the re-ID loss, including the classification loss and the proxy triplet loss. Note that we use the re-ID loss instead of regression loss for supervising the bounding boxes, and we only take one detected image as the example.}
	\label{fig:network}
\end{figure*}

\section{Methods}
We propose a novel localization refinement framework for person search task, which can be trained in an end-to-end fashion. The overview of our model is shown in Fig.~\ref{fig:network}. Traditional detection networks are generally supervised by two losses. A classification loss is used to distinguish the object categories or background. A regression loss is employed to adjust bounding boxes. 
However, for the person search task, there is no guarantee that the bounding box supervised by regression loss can produce the most suitable one for the downstream person re-ID task, as shown in Fig.~\ref{fig:motivation}. So we redesign the supervision of the detector in our person search framework. Specifically, we develop a differentiable ROI transform layer (Sec.~\ref{subsec:roi}) to realize the cropping operation of detected bounding boxes from original images. Then the cropped images are sent to a fixed re-ID model with the proxy triplet loss (Sec.~\ref{subsec:proxy}) and softmax loss. Since the whole process is differentiable w.r.t. the box coordinates, the detection network can be optimized by re-ID loss in an end-to-end framework, making the refined bounding boxes more reliable for the person search task. (Sec.~\ref{subsec:model}).

\subsection{ROI transform layer}\label{subsec:roi}
Followed by the detection network, we design a differentiable ROI transform layer to realize the cropping operation of bounding boxes from raw images. It aims at making the framework end-to-end trainable. There are two reasons to crop the images of bounding boxes. Firstly, although the corresponding features of bounding boxes are available in the general detection network, it leads to a shared feature embedding of the two tasks. This may cause conflicts~\cite{chen2018person}. Secondly, the receptive field is big enough in feature maps. Thus the cropping operation may introduce redundant context, which is harmful to re-ID. Therefore, it is superior to crop the images of bounding boxes.
\begin{figure}[htbp]
	\begin{center}
		\includegraphics[width=0.8\linewidth]{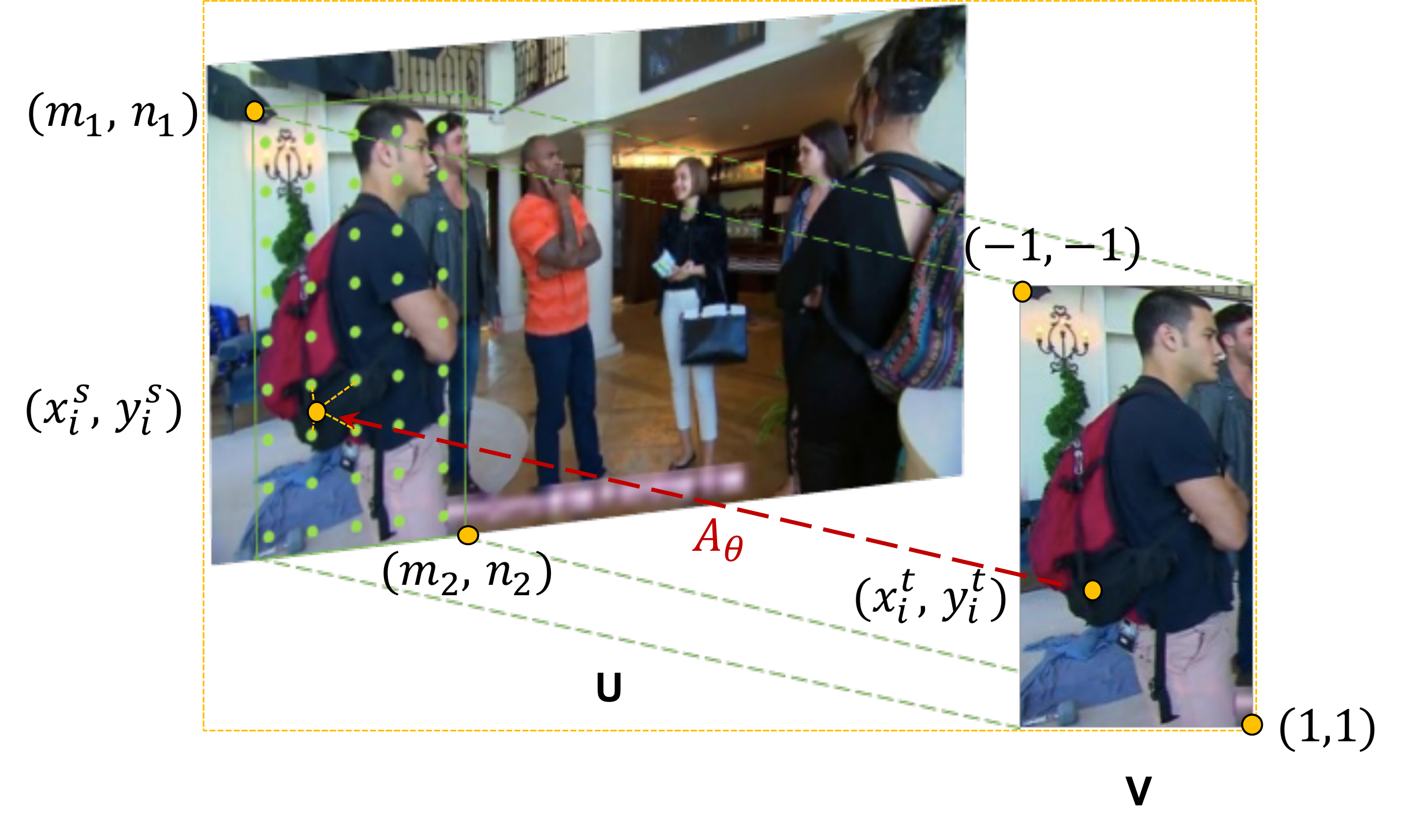}
	\end{center}
	\caption{Illustration of sampling grid from the source image $U$ to produce the target image $V$. }
	\label{fig:roi}
\end{figure}
Suppose the upper left and lower right coordinates of the bounding box are $(m_1,n_1,m_2,n_2)$ in the raw image. In order to transform the bounding box from the original image, as Fig.~\ref{fig:roi} shows, we first need to compute an affine transformation as follows:
\begin{equation}
\begin{pmatrix}
x_i^s\\y_i^s 
\end{pmatrix}
=A_{\theta}
\begin{pmatrix}
x_i^t \\ y_i^t \\ 1
\end{pmatrix} 
\label{eq:trans},
\end{equation}
where $(x_i^s, y_i^s)$ denote the source coordinates in the original image. $(x_i^t, y_i^t)$ denote the target coordinates of the regular grid in the output image with $H \times W$. We use normalized height and width target coordinates, so $x_i^t, y_i^t \in [-1,1] $. The corresponding coordinates are shown in Fig.~\ref{fig:roi}. $A_{\theta}$ is the affine transformation matrix, which is adaptively learned in the network in some methods~\cite{jaderberg2015spatial} due to the lack of supervised information. However, we have the source coordinates $(m_1,n_1,m_2,n_2)$ and the corresponding normalized target coordinates $(-1,-1,1,1)$. Therefore, the parameters of the transformation matrix can be calculated as follows:
\begin{equation}
A_{\theta}=\frac{1}{2}
\begin{bmatrix}
m_2-m_1&0 &m_2+m_1\\
0 &n_2-n_1&n_2+n_1
\end{bmatrix}.
\label{eq:theta}
\end{equation}

The grid of pixels on the target image $V$ is represented by $P^t=\{(x_i^t,y_i^t), i=1,2...N\}$, $N$ is the number of pixels. For each point on $V$, we compute the corresponding one on $U$ using Eq.~\ref{eq:trans}. So we can obtain the grid on $U$ denoted as $P^s=\{(x_i^s,y_i^s), i=1,2...N\}$. This process can back-propagate gradients since Eq.~\ref{eq:theta} is differentiable.

The pixel value of $(x_i^t,y_i^t)$ is bilinearly interpolated from the pixels near $(x_i^s,y_i^s)$ on the source image $U$. By setting all pixel values, we get the target image $V$:
\begin{equation}
V = B(P^s, U),
\end{equation}
where $B$ represents the bilinear sampler~\cite{jaderberg2015spatial}, which is also a differentiable module.

\subsection{Localization Refinement}\label{subsec:model}

Considering the optimization objective of a separated detector is not the same as the person search task, we optimize the localization of the detector with the supervision of re-ID loss in an end-to-end framework. In our method, we replace the regression loss with the re-ID loss, containing the proxy triplet loss and the softmax loss (in some cases). Based on the ROI transform layer, images of the bounding boxes can be transformed from the raw image. Thus the connection is established between the two tasks, 
allowing gradients to be backpropagated through from re-ID loss to the coordinates of bounding boxes.
 Consequently, localization of the detector can be refined, and it is consistent with the person search. 

Specifically, the detector and re-ID model in our framework are pretrained on the training set. Raw images are fed to the detection network for generating bounding boxes, which are input to the ROI transform layer for cropping, then we obtain the corresponding images. Following is a re-ID model with the parameters fixed. Thus the bounding boxes can be refined under the supervision of re-ID loss. The derivatives of re-ID loss with respect to the bounding box coordinates $(m_1,n_1,m_2,n_2)$ can be calculated. Here we take $m_1$ as the example and its derivative is as follows:
\begin{equation}	
\begin{aligned}
\frac{\partial L_{reid}}{\partial m_1}=\frac{\partial L_{reid}}{\partial I'} \cdot \frac{\partial I'}{\partial m_1} \\
\end{aligned},
\label{eq:losses}
\end{equation} 
where $I'$ denotes the image exported by the ROI transform layer. Since the derivatives of $I'$ with respect to the bounding box coordinate is differentiable as described in Sec.~\ref{subsec:roi}. Thus the Eq.~\ref{eq:losses} is totally derivable.

\subsection{Proxy Triplet Loss}\label{subsec:proxy}
To confirm the expandability of our framework, we also employ the re-ID model pretrained on other datasets. Thus the softmax is discarded because of the different identities. It is necessary to develop a metric loss for solving this problem. Recently, triplet loss with the hard mining method (TripHard loss) has been widely used in person re-ID~\cite{hermans2017defense,wang2018mancs, wang2018learning}, which can achieve superior performance.

However, TripHard loss is infeasible in the person search task for two factors. Firstly, in TripHard loss, the mini-batch is comprised of $N$ identities, and each identity has $K$ images, where $K$ and $N$ are usually set to $4$ and $8$, respectively. But for person search, only a few high-resolution raw images can be fed into the network for training, so the detected images may be insufficient. Secondly, the detected persons often do not have positive pairs such that it is infeasible to construct a triplet. Thus, we propose to build a triplet consisting of an anchor (the sample), a positive and a negative proxy selected through the proxy table. We call the method as the proxy triplet loss, described as follows. 
\begin{figure}[htbp]
	\begin{center}
		\includegraphics[width=0.8\linewidth]{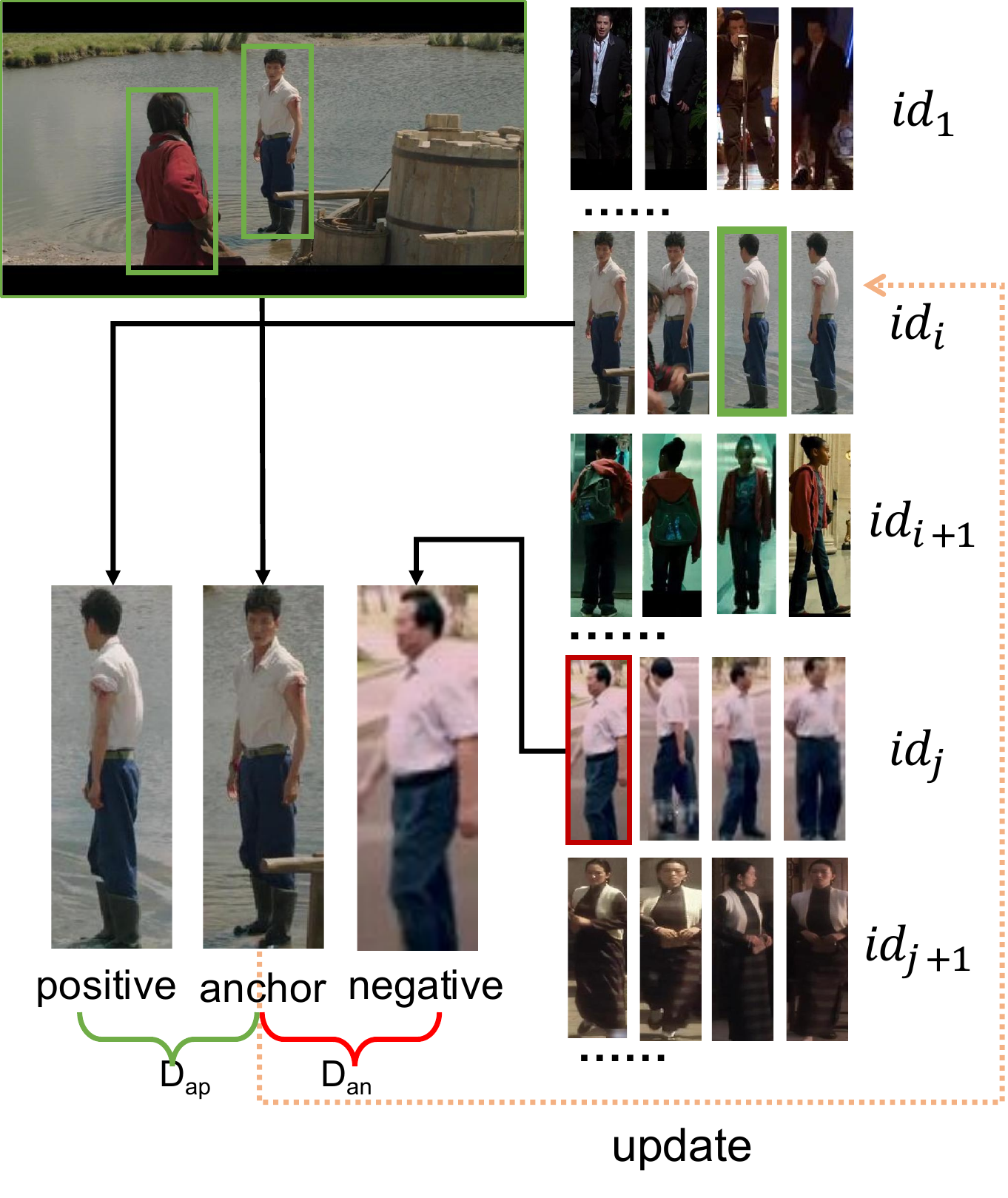}
	\end{center}
	\caption{Diagram of the proxy triplet loss. In a mini-batch, all the detected samples are seen as the anchor and we take one for example. Suppose the identity is $\textrm{ID}_i$. Thus the $i$ row of the table stores the positive proxies while the rests are negative ones. Then we can select the hardest positive and the hardest negative proxy to build a triplet. During backward, the samples are used to update the positive proxy by the first-in-first-out rule.}
	\label{fig:proxy}
\end{figure}
To construct triplet at each iteration, we need to assign proxies for each sample. As seen in Fig.~\ref{fig:proxy}, we build a proxy table $T \in \mathbb{R}^{N \times K}$ to store the proxy features. $N$ and $K$ represent the number of labeled identities and the volume of each identity, respectively. The features are filled with zeros at initialization. During the forward propagation, the proxy triplet loss is computed, while in backward, the table is updated with the samples. Specifically, we use the latest sample features to replace the corresponding proxy features with the same identity by the first-in-first-out rule. 

Based on the proxy table, we consider assigning the proxies for each sample to build triplet at each iteration. Specifically, suppose there are $b$ detected samples $I_i,i=1...b$, and each one is seen as the anchor. 
As shown in Fig~\ref{fig:proxy}, for a specified sample, we select the positive proxies from the corresponding identity in the table, and the rest are the negative proxies. By employing hard mining, we choose the hardest positive and the hardest negative proxy to build the triplet. The proxy triplet loss can be computed as follows:
\vspace{-3mm}
\begin{equation}
\begin{aligned}
L _{tri}=\sum_{i=1}^{b} [m&+\max_{p=1...K}D(f(I_i),f(I_i^p))\\
&-\min_{\substack{ j=1...b \\ n=1...K  \\ j\ne i}}D(f(I_i),f(I_j^n))]_+
\end{aligned},
\end{equation} 
where $f(\cdot )$ represents the feature embedding output from the identification network, $D(\cdot )$ measures the squared Euclidean distance between the two features. $m$ controls the margin between negative sample pairs and positive sample pairs.

\begin{table*}[htbp]
	\small
	\begin{center}
		\caption{Data statistics and evaluation setting of the CUHK-SYSU and PRW datasets. Bbox: Bounding box.}
		\label{tab:dataset}
		\vspace{1mm}
		\begin{tabular}{|c|ccc|ccc|ccc|}
			\hline
			\multirow{2}*{Methods} & \multicolumn{3}{|c|}{all} & \multicolumn{3}{|c}{train}&\multicolumn{3}{|c|}{test}\\
			\cline{2-10}
			~&Images&Bboxes&IDs&Images&Bboxes&IDs&Images&Bboxes&IDs\\
			\hline			
			\hline
			CUHK-SYSU~\cite{xiao2017joint} & 18184 & 96143 & 8432 &11206 & 55272 & 5532 & 6978 & 40871 & 2900 \\			
			PRW~\cite{zheng2017person} & 11816 & 43110 & 932 & 5704 & 18048 & 482 & 6112 & 25062 & 450\\
			\hline
		\end{tabular}
	\end{center}
\vspace{-4mm}
\end{table*}

\section{Experiment}
In this section, we first describe two large datasets and the evaluation protocols, as well as some implementation details. Then we compare our method with state-of-the-art methods.
Afterward, we conduct several ablation studies to explore the effects of different components. 

\subsection{Datasets and Protocols}
\paragraph{CUHK-SYSU.} CUHK-SYSU~\cite{xiao2017joint} is a large scale person search dataset captured by a hand-held camera in the street/urban scene or chosen from the movie snapshots, which is consisted of $18,184$ images and $96,143$ annotated bounding boxes. There are $8,432$ labeled identities, and the rest of the annotated ones are considered as the negative samples. 
The training set includes $11,206$ images and $5,532$ identities, while the testing set contains $6,978$ gallery images and $2,900$ probe images. 
\vspace{-3mm}
\paragraph{PRW.} PRW~\cite{zheng2017person} dataset
 is captured by six spatially disjoint cameras. There are 
 $11,816$ frames with the $43,110$ annotated bounding boxes. The identifications are ranged from $1$ to $932$, and the rest are viewed as the unknown persons with the identification of $-2$. 
The training set contains $5,704$ frames and $482$ identities, while the testing set includes $6,112$ gallery images and $2,057$ query images with $450$ identities.
\vspace{-3mm}
\paragraph{Evaluation Protocols.} In our experiments, we adopt the evaluation metrics following~\cite{xiao2017joint}, namely the cumulative matching cure (CMC) and the mean Average Precision (mAP). The first one is widely used in person re-ID. The second one is inspired by the object detection task. We use the ILSVRC object detection criterion~\cite{russakovsky2015imagenet} to measure the correctness of the predicted bounding boxes. Based on the precision-recall curve, an averaged precision (AP) is calculated for each query, then average the APs of all the queries to get the final result.
\begin{table}[htbp]
	\footnotesize
	\begin{center}
		\caption{Experimental comparisons with state-of-the-art methods on CUHK-SYSU.}
		\label{tab:sysu}
		\vspace{1mm}
		\begin{tabular}{|l|c|c|}
			\hline
			Method&Rank-1(\%)&mAP(\%) \\
			\hline			
			\hline
			ACF~\cite{dollar2014fast}+DSIFT~\cite{zhao2013unsupervised}+Euclidean & 25.9 & 21.7 \\			
			ACF~\cite{dollar2014fast}+DSIFT~\cite{zhao2013unsupervised}+KISSME~\cite{koestinger2012large} & 38.1& 32.3  \\
			ACF~\cite{dollar2014fast}+LOMO+XQDA~\cite{liao2015person} & 63.1 & 55.5  \\
			\hline	
			CCF~\cite{yang2015convolutional}+DSIFT~\cite{zhao2013unsupervised}+Euclidean & 11.7 & 11.3 \\			
			CCF~\cite{yang2015convolutional}+DSIFT~\cite{zhao2013unsupervised}+KISSME~\cite{koestinger2012large} & 13.9 & 13.4  \\
			CCF~\cite{yang2015convolutional}+LOMO+XQDA~\cite{liao2015person} & 46.4 & 41.2  \\
			CCF~\cite{yang2015convolutional}+IDNet & 57.1 & 50.9  \\
			\hline
			CNN~\cite{ren2015faster}+DSIFT~\cite{zhao2013unsupervised}+Euclidean & 39.4 & 34.5 \\			
			CNN~\cite{ren2015faster}+DSIFT~\cite{zhao2013unsupervised}+KISSME~\cite{koestinger2012large} & 53.6 & 47.8  \\
			CNN~\cite{ren2015faster}+Bow~\cite{zheng2015scalable}+Cosine & 62.3 & 56.9  \\				
			CNN~\cite{ren2015faster}+LOMO+XQDA~\cite{liao2015person} & 74.1 & 68.9  \\
			CNN~\cite{ren2015faster}+IDNet & 74.8 & 68.6  \\			
			\hline
			OIM~\cite{xiao2017joint} & 78.7 & 75.5  \\
			NPSM~\cite{liu2017neural} & 81.2 & 77.9  \\
			RCAA~\cite{chang2018rcaa} & 81.3 & 79.3 \\
			I-NeT~\cite{he2018end} &81.5 & 79.5 \\
			MGTS~\cite{chen2018person} & 83.7 & 83.0  \\
			CLSA~\cite{lan2018person} & 88.5 & 87.2  \\
			\hline
			Ours &\textbf{94.2} & \textbf{93.0}  \\
			\hline
		\end{tabular}
	\end{center}
\vspace{-2mm}
\end{table}
\begin{table}[htbp]
	\footnotesize
	\begin{center}
		\caption{Experimental comparisons with state-of-the-art methods on PRW.}
		\label{tab:prw}
		\vspace{1mm}
		\begin{tabular}{|l|c|c|}
			\hline
			Method&Rank-1(\%)&mAP(\%) \\
			\hline			
			\hline
			ACF-Alex~\cite{dollar2014fast}+LOMO+XQDA~\cite{liao2015person} & 30.6 & 10.3  \\
			ACF-Alex~\cite{dollar2014fast}+IDE$_{det}$~\cite{zheng2017person} & 43.6 & 17.5 \\
			ACF-Alex~\cite{dollar2014fast}+IDE$_{det}$~\cite{zheng2017person}+CWS~\cite{zheng2017person} & 45.2 & 17.8 \\
			\hline	
			DPM-Alex~\cite{felzenszwalb2010object}+LOMO+XQDA~\cite{liao2015person} & 34.1 & 13.0  \\
			DPM-Alex~\cite{felzenszwalb2010object}+IDE$_{det}$~\cite{zheng2017person} & 47.4 & 20.3 \\
			DPM-Alex~\cite{felzenszwalb2010object}+IDE$_{det}$~\cite{zheng2017person}+CWS~\cite{zheng2017person} & 48.3 & 20.5 \\
			\hline
			LDCF~\cite{nam2014local}+LOMO+XQDA~\cite{liao2015person} & 31.1 & 11.0  \\
			LDCF~\cite{nam2014local}+IDE$_{det}$~\cite{zheng2017person} & 44.6 & 18.3 \\
			LDCF~\cite{nam2014local}+IDE$_{det}$~\cite{zheng2017person}+CWS~\cite{zheng2017person} & 45.5 & 18.3 \\			
			\hline
			OIM~\cite{xiao2017joint} & 49.9 & 21.3  \\
			NPSM~\cite{liu2017neural} & 53.1 & 24.2  \\
			CLSA~\cite{lan2018person} & 65.0 & 38.7  \\
			MGTS~\cite{chen2018person} & \textbf{72.1}& 32.6 \\
			\hline
			Ours & 70.2 & \textbf{42.9}  \\
			\hline
		\end{tabular}
	\end{center}
\vspace{-5mm}
\end{table}

\subsection{Implementation Details}	
For the detection network, we use the latest PyTorch implementation of Faster R-CNN~\cite{ren2015faster} released by Facebook research~\footnote{\url{https://github.com/facebookresearch/Detectron}}. The backbone network is ResNet-50 with FPN. Our detector is pretrained on the SYSU and PRW with the mAP of $91.1\%$ and $94.9\%$, respectively. For the re-ID network, we adopt a popular baseline~\footnote{\url{https://github.com/L1aoXingyu/reid_baseline}} based on the Resnet-50. The results of the pretrained model on the SYSU and PRW datasets achieve $91.8\%$ and $76.2\%$ on Rank-1, respectively.
\begin{figure}[htbp]
	\begin{center}
		\includegraphics[width=0.8\linewidth]{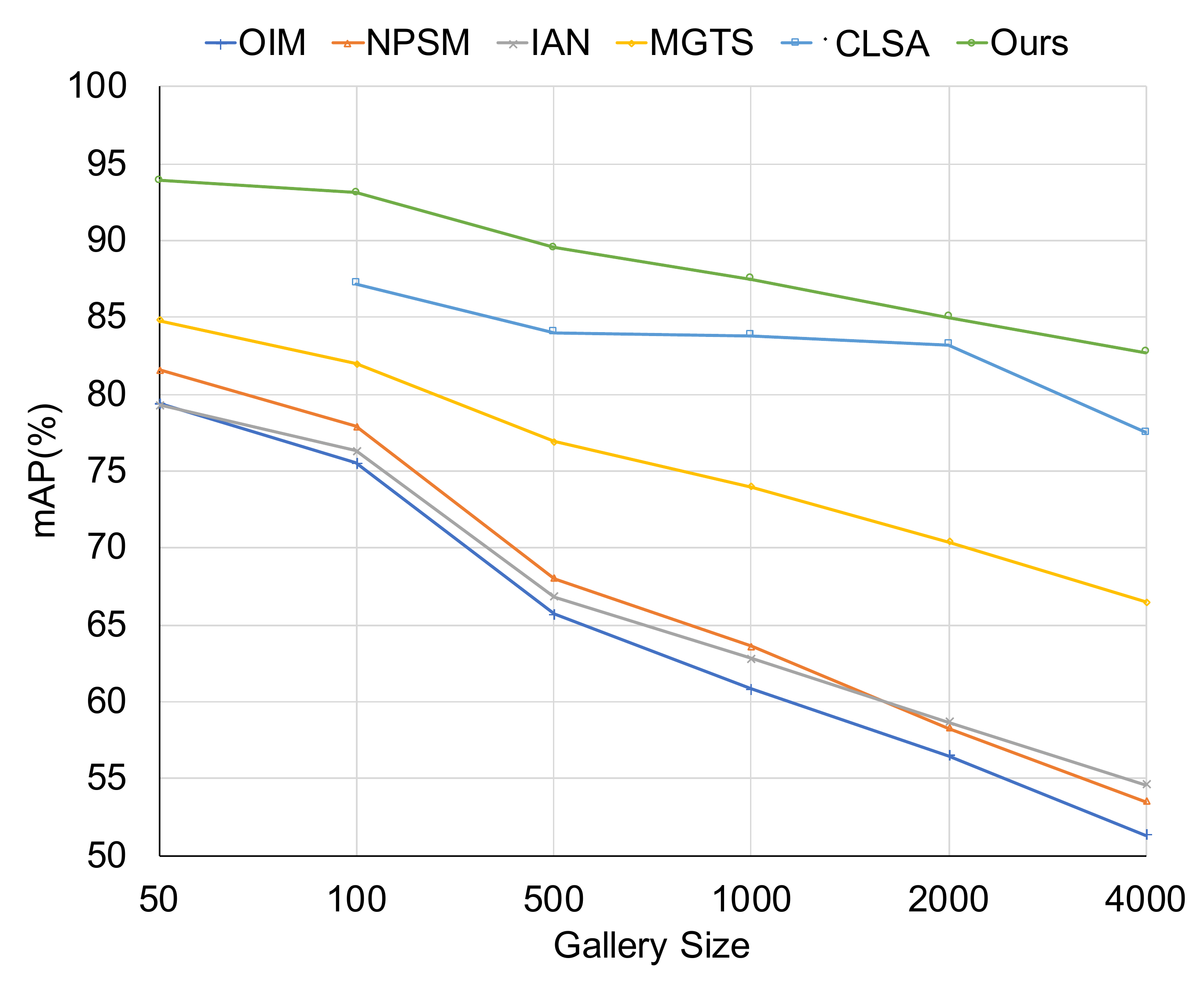}
	\end{center}
	\vspace{-3mm}
	\caption{Performance comparison on CUHK-SYSU with different gallery sizes. }
	\label{fig:gallery_size}
	\vspace{-3mm}
\end{figure}
For our end-to-end optimization framework, we remove the original regression loss and use the proxy triplet loss as well as softmax loss to supervise the whole network. The batch size is set to $4$ because of the limitation of GPU memory. The volume of the proxy triplet loss is set to $2$, so it can satisfy the triplet hard construction with little overhead. And we set the triplet margin to $0.5$. We choose the batched Stochastic Gradient Descent (SGD) optimizer with the momentum of $0.9$. The weight decay factor for L2 regularization is set to $5\times10^{-4}$. As for the learning rate strategy, we use the warm-up schedule. The base learning rate is $0$, which warms up to $5\times10^{-5}$ in the first 500 iterations, then decays to $5\times10^{-6}$ after $1\times10^{4}$ iterations. Our model is trained for $4\times10^{4}$ iterations totally.
All experiments are conducted on the publicly available PyTorch platform, and the network is trained on $4$ NVIDIA TITAN XP GPUs.

When evaluating the influence of gallery size on the CUHK-SYSU dataset, we use a varying gallery size of $\{50, 100, 500, 1000, 2000, 4000\}$. In the following experiment, the results are represented with the gallery size of 100 by default. 

\subsection{Comparisons with the state-of-the-art methods}
We compare our proposed network with current state-of-the-art methods including OIM~\cite{xiao2017joint}, NPSM~\cite{liu2017neural}, RCAA~\cite{chang2018rcaa}, I-Net~\cite{he2018end}, MGTS~\cite{chen2018person}, CLSA~\cite{lan2018person} on two popular datasets CUHK-SYSU~\cite{xiao2017joint} and PRW~\cite{zheng2017person}. In addition to these methods, we also compare with the methods that joint different pedestrian detectors (DPM~\cite{felzenszwalb2010object}, ACF~\cite{dollar2014fast}, CCF~\cite{yang2015convolutional}, LDCF~\cite{nam2014local}, and R-CNN~\cite{girshick2014rich}) and person descriptors (BoW~\cite{zheng2015scalable}, LOMO~\cite{liao2015person}, DenseSIFT-ColorHist (DSIFT)~\cite{zhao2013unsupervised}) and distance metric (KISSME~\cite{koestinger2012large}, XQDA~\cite{liao2015person}).

\vspace{-3mm}
\paragraph{Evaluation On CUHK-SYSU.}
Tab.~\ref{tab:sysu} shows the performance comparison between our network and existing competitive methods on the CUHK-SYSU dataset. The gallery size is set to 100. We follow the notations definition in~\cite{xiao2017joint}, where ``CNN" and ``IDNet" represent the Faster R-CNN detector based on ResNet-50 and the re-ID net, respectively. It can be seen that our method significantly outperforms all other competitors, including both the end-to-end and the separated approach. Compared with the state-of-the-art method CLSA, our method obtains the performance gain of 5.8\%/5.7\% in terms of Rank-1/mAP metric.

In order to evaluate the performance scalability of our model, we compare with other competitive methods under different gallery sizes. As Fig.~\ref{fig:gallery_size} shows, we evaluate the mAP with a varying gallery size of $[50, 100, 500, 1000, 2000, 4000]$. We can see that except CLSA all methods degrade the performance as the gallery size increases. Nevertheless, our approach stands out with great advantages under different settings, which indicates the robustness of our model. When the gallery size increases from 50 to 4000, the declining extent of our method is relatively small compared with others, which shows the scalability when dealing with large scale person search problem. With the increasing scales, more distracting people are involved in the identity matching process, which is close to real-world applications. This indicates the importance of refining the bounding boxes.

\vspace{-3mm}
\paragraph{Evaluation On PRW.}
We further evaluate our method with the competitive ones on the PRW dataset, and the results are shown in Tab.~\ref{tab:prw}. Note that under the benchmarking setting~\cite{zheng2017person}, the gallery contains all the 6112 testing images. 
It can be seen that our method outperforms the CLSA by 5.2\%/4.2\% on Rank-1 and mAP. Compared with MGTS, we gain the promotion of 10.3\% on mAP. This confirms the effectiveness of our method in the person search task.
\begin{figure*}[htbp]%
	\setlength{\abovecaptionskip}{0pt} 
	\centering
	\subfigure[Rank-1 on same dataset]{%
		\label{fig:size_1}%
		\includegraphics[width=1.75in]{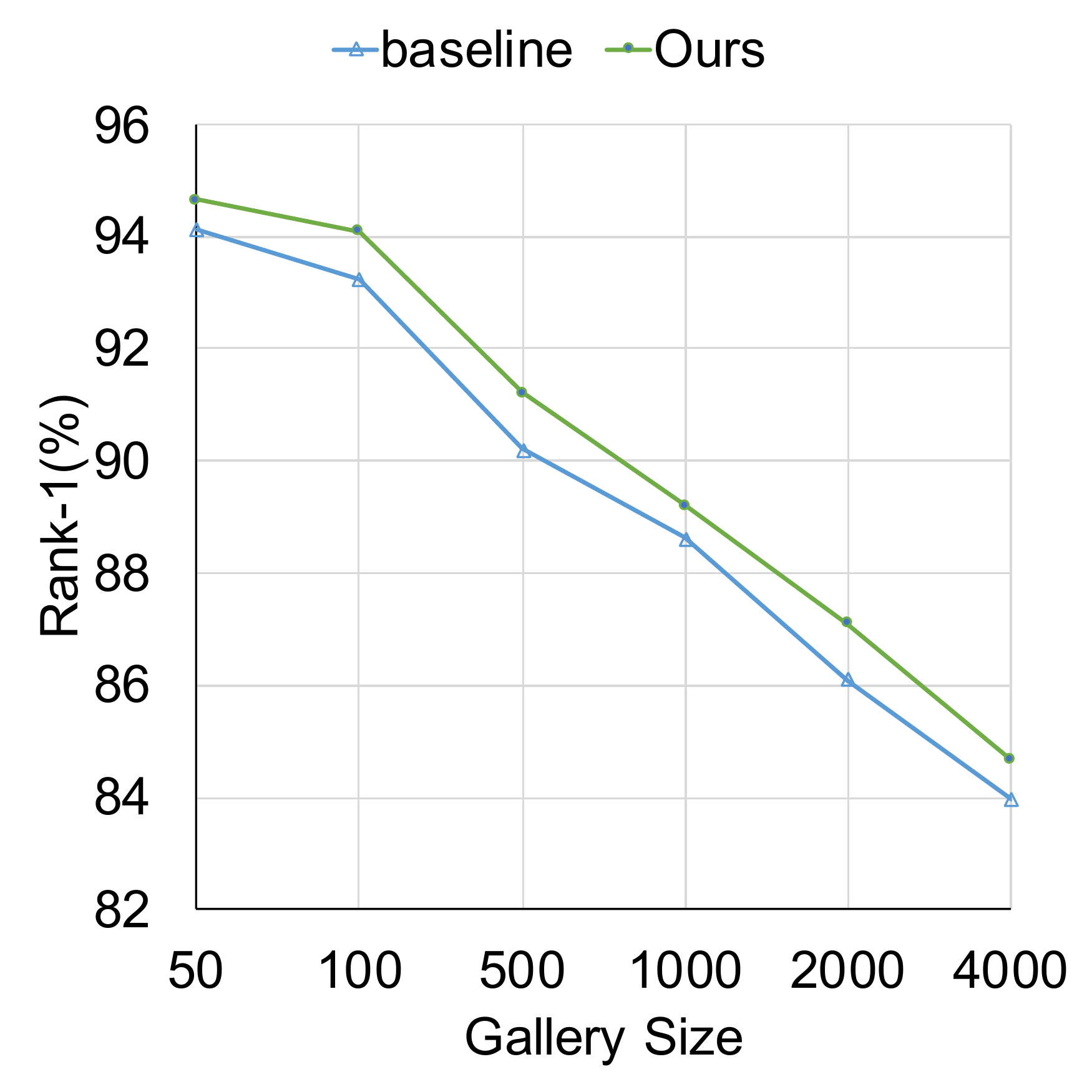}}%
	\subfigure[mAP on same dataset ]{%
		\label{fig:size_2}%
		\includegraphics[width=1.75in]{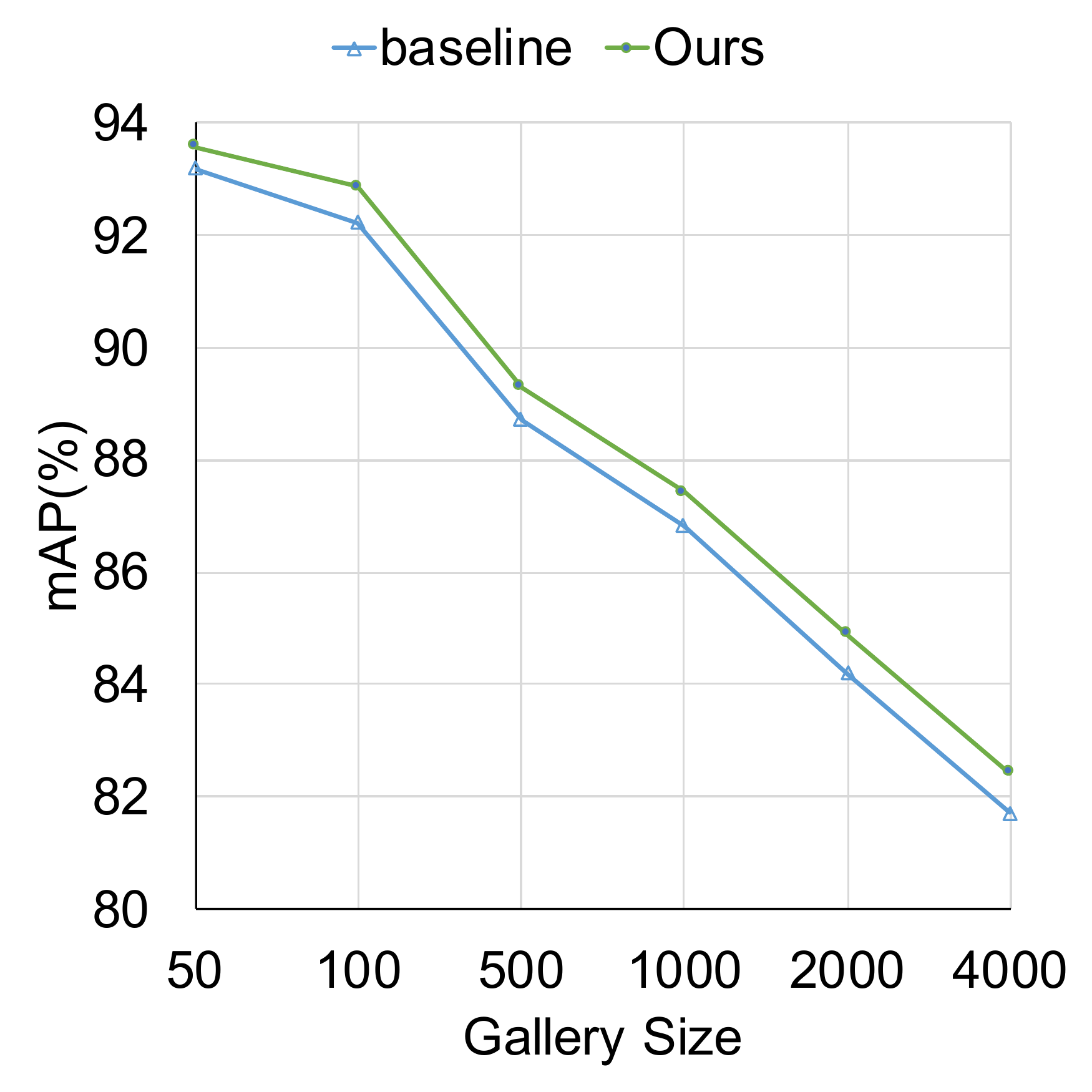}}%
	\subfigure[Rank-1 on cross dataset]{%
		\label{fig:size_3}%
		\includegraphics[width=1.75in]{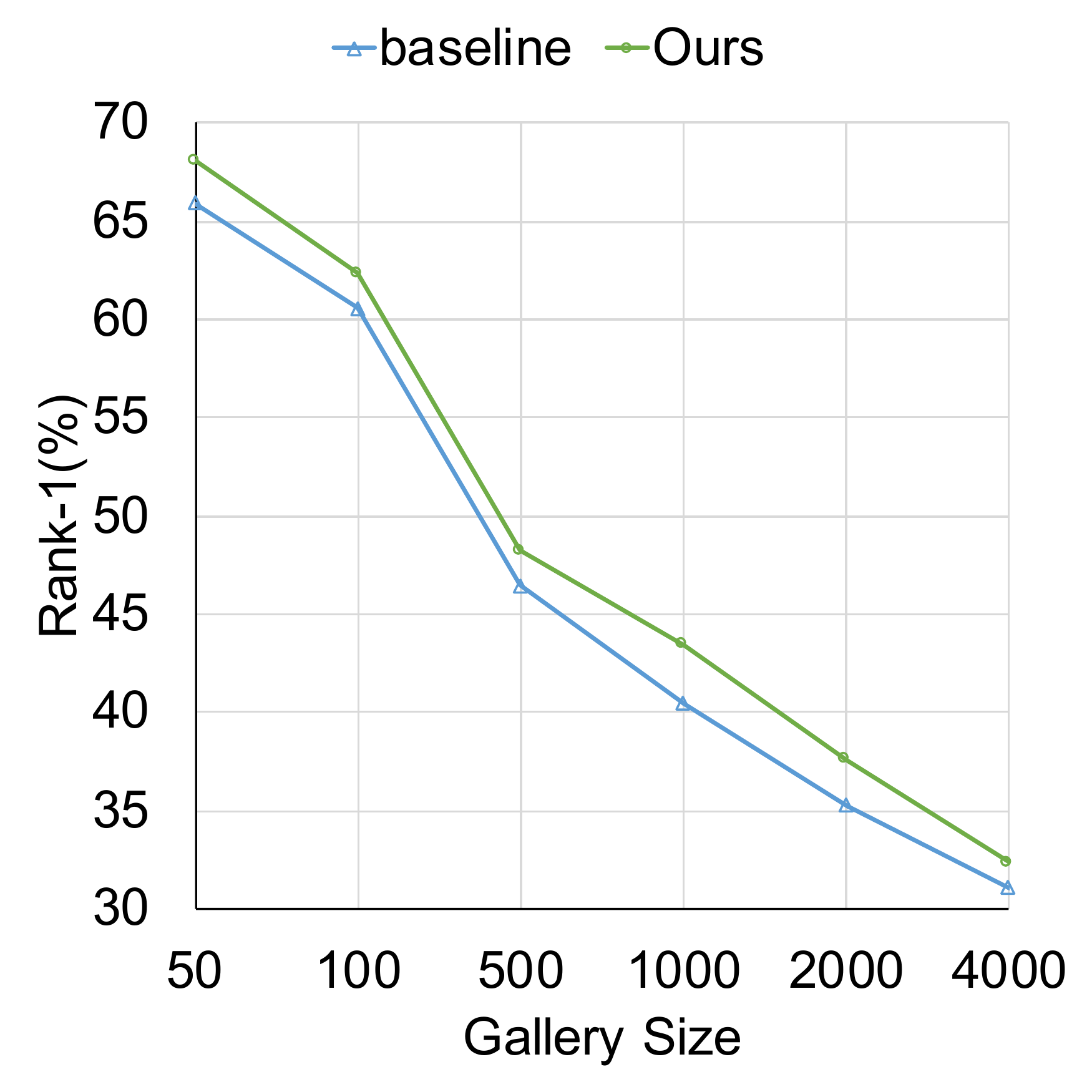}}%
	\subfigure[mAP on cross dataset]{%
		\label{fig:size_4}%
		\includegraphics[width=1.75in]{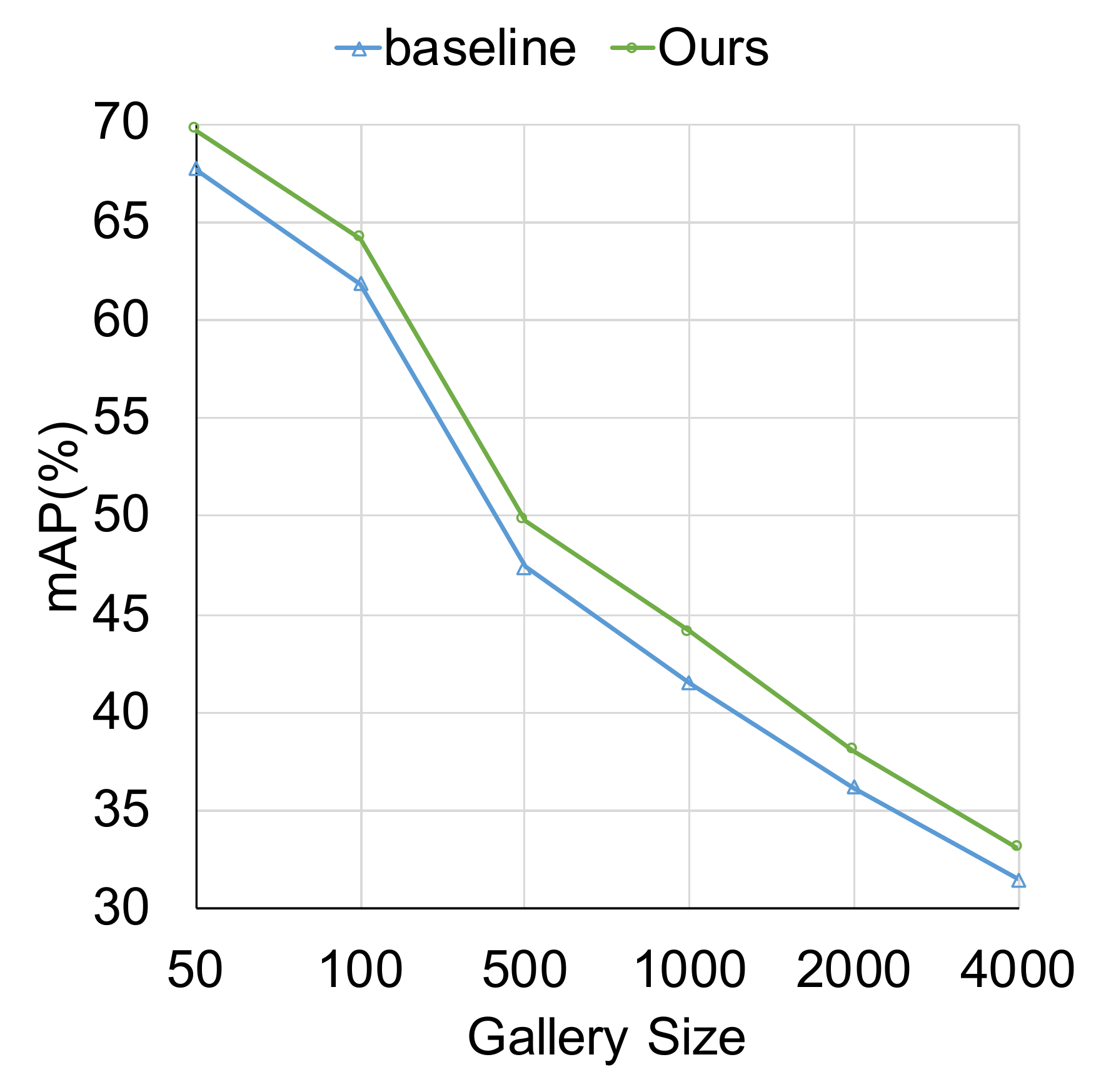}}%
	\caption{Performance comparisons between baseline and our method on CUHK-SYSU with different gallery sizes. (a)(b) are conducted on the same dataset while (c)(d) are conducted on the cross datasets. The difference is that the re-ID model is pretrained on the CUHK-SYSU and Market-1501 dataset, respectively. The evaluation protocol is the CMC Rank-1 and mAP. }
	\label{fig:ablation_size}
\end{figure*}

\subsection{Ablation study}
In this section, we first show the promotion on different detectors with the refinement of our method. Second, we analyze the results with the re-ID model trained on the same dataset and cross dataset, respectively. Third, we show the effectiveness of our proposed proxy triplet loss. The experiments are conducted on the CUHK-SYSU dataset.

\begin{table}[htbp]
	\small
	\begin{center}
		\caption{The results on CUHK-SYSU with different detectors.}
		\label{tab:ablation_detector}
		\vspace{1mm}
		\begin{tabular}{|c|c|c|}
			\hline
			Method&Rank-1(\%)&mAP(\%) \\
			\hline
			\hline
			Faster R-CNN & 93.24 & 92.23  \\
			Faster R-CNN +ours & 94.24 & 93.09  \\
			\hline
			RetinaNet & 92.21 & 91.84 \\
			RetinaNet +ours & 93.97 & 92.78  \\
			\hline
		\end{tabular}
	\end{center}
	\vspace{-7mm}
\end{table}

\vspace{-3mm}
\paragraph{Different detectors.}
In order to evaluate the expandability of our method, we combine different detection network in our model, including Faster R-CNN~\cite{ren2015faster} and RetinaNet~\cite{lin2017focal}. The results are shown in Tab.~\ref{tab:ablation_detector}, When the detector is trained separately, the results on re-ID have achieved the state-of-the-art. Then we supervise the bounding boxes with the re-ID loss instead of the original regression loss. After the refinement by our method, the performance is improved again by 1.0\%/1.76\% on Rank-1 and 0.86\%/0.94\% on mAP with Faster R-CNN and RetinaNet, respectively. The results confirm that our approach is robust and can be extended to multiple detectors.

\begin{figure*}[htbp]%
	\begin{center}
		\includegraphics[width=0.84\linewidth]{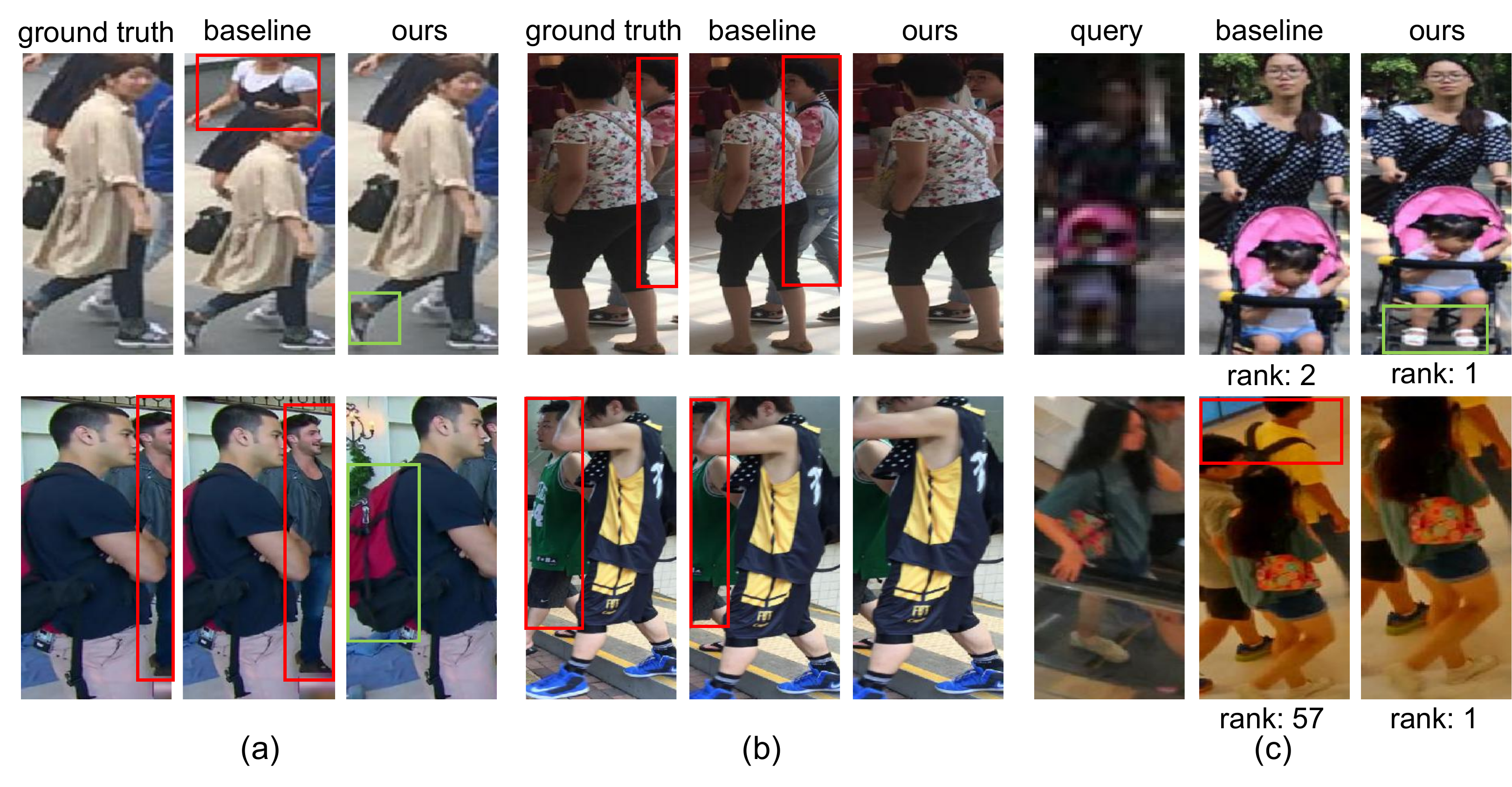}
	\end{center}
	\caption{Comparisons of the individually trained detector and our method. In (a)(b), the ground truth is manually labeled, and the baseline is detected by the individually trained detector while the last is refined by our methods. In (c), the first is the query; the second is the matching result with the individual detector, and the last is the result of our methods. We omit the ranking lists, and the ranks are shown below the images.}
	\label{fig:ablation_compare}
	\vspace{-3mm}
\end{figure*}

\vspace{-3mm}
\paragraph{Same dataset and cross dataset.}
The detector is pretrained on CUHK-SYSU, while the re-ID model is pretrained on CUHK-SYSU and Market-1501 for the same dataset and cross dataset person search, respectively. The baseline adopts the individually trained detector to produce bounding box images, then input to the trained re-ID model for testing. Our method combines the two models in a framework, optimizing the detector with the fixed re-ID model. Results with different gallery size are shown in Fig.~\ref{fig:ablation_size}. For the same dataset, We can observe that our strong baseline surpasses the state-of-the-art methods. Moreover, our method outperforms baseline with all gallery size, especially in a large gallery. The Rank-1 and mAP are shown in Fig.~\ref{fig:size_1} and Fig.~\ref{fig:size_2}. The results confirm our method is effective on the large scale dataset. For cross dataset, the results are shown in Fig.~\ref{fig:size_3} and Fig.~\ref{fig:size_4}. The base Rank-1/mAP is 60.59\%/61.87\%, after the optimizing of our method, the results can be promoted to 62.34\%/64.15\% with the gallery size of 100. 
\vspace{-3mm}
\paragraph{Visualization and Analysis.}
To exhibit the effect of our method intuitively, a visualization in testing is illustrated in Fig.~\ref{fig:ablation_compare}. In (a)(b), the baseline is detected by the individually trained detector. From (a) we can see that our method can remove the interferential person compared with the individual detector. And some attributes like shoes and bags are included, which even is superior to the ground truth. In (b), the excessive background in ground truth can be removed by our methods. In (c), given the query, it is seen that the matching results of ours are superior to the baseline because the refined detector is more reliable for person search task.

\vspace{-3mm}
\paragraph{Effect of the different loss.}
We conduct extensive experiments on different losses, note that the classification loss is in the re-ID model rather than the one in the detection network. The baseline denotes the individually trained detector and re-ID model. Others are based on our method, using the re-ID loss other than regression loss to supervise the detector. As can be seen from Tab.~\ref{tab:ablation_loss}, all of our methods can outperform the baseline. Moreover, We can obtain a superior performance of 94.24\%/93.09\% on Rank-1/mAP when combining the two losses together. This confirms the effectiveness of our proxy triplet.
\begin{table}[htbp]
	\small
	\begin{center}
		\caption{Supervised with different losses. $L_{cls}$ and $L_{tri}$ denote the classification loss and our proposed proxy triplet loss, respectively.}
		\label{tab:ablation_loss}
		\vspace{1mm}
		\begin{tabular}{|c|c|c|}
			\hline
			Methods &Rank-1(\%)&mAP(\%) \\
			\hline
			\hline
			Baseline& 93.24 & 92.23  \\
			Ours($L_{cls}$)& 93.52 & 92.34  \\
			Ours($L_{tri}$)& 93.48 & 92.49  \\
			Ours($L_{cls}$+$L_{tri}$)& 94.24 & 93.09 \\
			\hline
		\end{tabular}
	\end{center}
\vspace{-6mm}		
\end{table}

\section{Conclusion}	
In this paper, we propose a localization refinement network that joints the task of pedestrian detection and person re-ID in an end-to-end framework. To solve the problem that the individual detector generates a sub-optimal bounding box for person re-ID, we optimize the detector with the supervision of re-ID loss to produce a reliable bounding box for person search. Specifically, we develop a differentiable ROI transform layer to conduct the cropping operation of the detected bounding boxes. Then the cropped images are fed to the re-ID model with the supervision of re-ID loss. Since the whole process is differentiable w.r.t. the box coordinates, the detection network can be optimized by re-ID loss in an end-to-end framework. Moreover, we design a proxy triplet loss to solve the problem that the standard triplet hard loss cannot be conducted in person search pipeline. Extensive comparative evaluations have been conducted on two large scale person search datasets CUHK-SYSU and PRW. The results validate the performance superiority of our method.	
\section*{Acknowledgements}	
This work was supported by the Project of the National Natural Science Foundation of China No.61876210, the Fundamental Research Funds for the Central Universities No.2019kfyXKJC024, and the 111 Project on Computational Intelligence and Intelligent Control under Grant B18024.
\clearpage
{\small
	\bibliographystyle{ieee_fullname}
	\bibliography{egpaper_final}
}

\end{document}